# A Cloud-Based Cross-Modal Transformer for Emotion Recognition and Adaptive Human–Computer Interaction

Ziwen Zhong[1], Zhitao Shu[2], Yue Zhao[3]

[1] Beijing University of Post & Telecommunication, Beijing, China
[2] Vanderbilt University, Nashville, TN, USA
[3] King graduate school, Monroe University, New Rochelle, NY, USA, 10801

[1] 3345258828@qq.com
[2] zhitao.shu@vanderbilt.edu
[3] Yz3160@nyu.edu

**Abstract.** Emotion recognition is a fundamental component of next-generation human-computer interaction (HCI), enabling machines to perceive, understand, and respond to users' affective states. However, existing systems often rely on single-modality analysis-such as facial expressions, speech tone, or textual sentiment-resulting in limited robustness and poor generalization in real-world environments. To address these challenges, this study proposes a Cloud-Based Cross-Modal Transformer (CMT) framework for multimodal emotion recognition and adaptive human-computer interaction. The proposed model integrates visual, auditory, and textual signals using pretrained encoders-Vision Transformer (ViT), Wav2Vec2, and BERT-and employs a cross-modal attention mechanism to capture complex interdependencies among heterogeneous features. By leveraging cloud computing infrastructure with distributed training (Kubernetes and TensorFlow Serving), the system enables scalable, low-latency emotion recognition for large-scale user interactions. Experiments conducted on benchmark datasets including IEMOCAP, MELD, and AffectNet demonstrate that the proposed Cloud-Based Cross-Modal Transformer (CMT) achieves state-of-the-art performance, improving the F1-score by 3.0% and reducing cross-entropy loss by 12.9% compared to the best fusion-based baselines. Additionally, cloud deployment evaluations show an average response latency of 128 ms, representing a 35% reduction compared with conventional transformer-based fusion systems. These results confirm that the proposed CMT framework enables efficient, real-time emotion recognition and adaptive feedback in interactive applications such as intelligent customer service, virtual tutoring, and affective computing interfaces.These results highlight the effectiveness and scalability of the proposed CMT framework, marking an important step toward cloud-native affective computing and emotionally intelligent interactive systems.

**Keywords:** Multimodal Emotion Recognition; Cross-Modal Transformer; Cloud Computing; Human-Computer Interaction; ViT; Wav2Vec2; BERT; Affective Computing.

# 1. Introduction

With the rapid advancement of artificial intelligence and cloud computing, emotion-aware human–computer interaction (HCI) has become a key research frontier in achieving intelligent, personalized, and adaptive systems [1]. Emotion recognition allows machines to interpret human affective states through multimodal cues—such as facial expressions, vocal intonation, and linguistic semantics—thereby enhancing interaction naturalness and user satisfaction [2,3]. However, existing emotion recognition systems often rely on single-modality features, resulting in insufficient robustness, weak contextual understanding, and degraded performance in complex real-world environments. Moreover, traditional local computing architectures struggle to handle the high computational demands and data heterogeneity inherent in multimodal emotion analysis, making scalability and real-time processing major challenges.

To address these limitations, this paper proposes a Cloud-Based Cross-Modal Transformer (CMT) framework designed for multimodal emotion recognition and adaptive human–computer interaction. The proposed CMT integrates visual, auditory, and textual modalities using pretrained backbones—Vision Transformer (ViT) for visual embeddings, Wav2Vec2 for speech representation, and BERT for textual semantics. A cross-modal attention mechanism is employed to dynamically capture interdependencies and align heterogeneous features across modalities. The cloud-based architecture leverages distributed computing frameworks such as Kubernetes and TensorFlow Serving, ensuring scalable deployment, parallel model training, and real-time inference with low latency.

The CMT framework is particularly suited for emotion-driven applications in virtual assistants, online learning environments, healthcare support systems, and intelligent customer service. By running on cloud infrastructure, it enables seamless integration with large-scale user data streams and adaptive feedback loops, making interactions context-aware and emotionally intelligent.

The main contributions of this work are summarized as follows:

(1) We propose a novel cloud-native cross-modal transformer that unifies visual, auditory, and textual modalities for robust and scalable emotion recognition.
(2) We design a cross-modal attention fusion mechanism to effectively align heterogeneous features and enhance emotion representation.
(3) We implement a cloud-based distributed architecture that enables real-time adaptive human–computer interaction.
(4) Extensive experiments conducted on benchmark datasets (IEMOCAP, MELD, and AffectNet) demonstrate the superior performance and scalability of the proposed method compared to existing multimodal fusion frameworks.

# 2. Related Work

Emotion recognition has been an essential research topic in affective computing, aiming to bridge human emotions and intelligent systems through the integration of multimodal information. Early studies primarily relied on single-modality analysis, such as facial expression recognition using convolutional neural networks (CNNs) [4] or speech emotion detection through handcrafted acoustic features and recurrent models [5]. Although these approaches achieved promising performance in constrained environments, they often failed to generalize across users, contexts, and modalities due to the lack of cross-modal correlation modeling.

## 2.1 Multimodal Emotion Recognition

To overcome the limitations of unimodal systems, recent research has shifted toward multimodal emotion recognition (MER), integrating information from visual, auditory, and textual streams. Fusion strategies can be broadly categorized into three levels: early fusion (feature-level), late fusion (decision-level), and hybrid or hierarchical fusion. Early fusion approaches combine raw or low-level features across modalities but suffer from synchronization and noise issues. In contrast, late fusion combines modality-specific predictions, which can neglect intermodal dependencies. Hybrid fusion, often implemented via attention mechanisms or graph neural networks, has shown superior performance by balancing modality-specific learning and cross-modal correlation. Representative datasets such as IEMOCAP [6], MELD [7], and AffectNet [8] have facilitated benchmarking and model evaluation across modalities.

*2.2 Transformer-Based Multimodal Learning*

Qin et al [9]. identify limitations of the two-step ERC paradigm that first extracts static features and then models dialogue structure. They propose integrating context and structure during fine-tuning, yielding BERT-ERC with prompt text, fine-grained classification and two-stage training, which significantly outperforms the old paradigm across four datasets and adapts to resource-limited and online scenarios.

Bousaid et al [10]. review CNN/GCN/ViT advances in facial emotion recognition, identify data scarcity as a bottleneck, and propose fine-tuning ViT-B16 with VGG16 via transfer learning, achieving the highest accuracy on RAF-DB and demonstrating the effectiveness and practicality of the ViT-CNN hybrid paradigm for facial affect analysis.

With the rise of Transformer architectures, cross-modal learning has gained significant attention. Models such as ViLBERT, UNITER, and CLIP have demonstrated the effectiveness of attention-based mechanisms in aligning heterogeneous modalities. In emotion recognition, multimodal transformers leverage self-attention and cross-attention to dynamically capture temporal and contextual relationships across modalities [11]. For instance, multimodal transformers combining BERT for textual understanding, ViT for visual processing, and Wav2Vec2 for acoustic representation have shown notable improvements in emotion classification accuracy and robustness. However, most existing models are designed for offline processing and local computation, limiting their scalability and real-time adaptability in cloud-based interactive systems.

*2.3 Adaptive Human‐Computer Interaction (HCI)*

In parallel, the field of adaptive HCI has evolved toward emotion-driven personalization. Emotion-aware interfaces can adjust tone, color, dialogue style, and system behavior according to users' affective states [12]. Studies in social robotics, virtual assistants, and online education demonstrate that emotional intelligence significantly enhances user engagement and satisfaction. However, the majority of emotion-adaptive systems depend on pre-trained local models or rule-based adaptation strategies, which lack scalability, real-time feedback, and cross-domain transferability.

*2.4 Cloud Computing and Distributed Affective Systems*

Recent advances in cloud computing have enabled large-scale deployment of deep learning models for real-time multimodal processing. Frameworks such as Kubernetes [13], TensorFlow Serving, and AWS SageMaker provide distributed computation, elasticity, and fault tolerance for continuous learning and adaptation. Cloud-based architectures also facilitate federated emotion learning, where user data from different devices are securely integrated for model optimization while preserving privacy. Despite these advantages, few works have explored the integration of cross-modal transformers with cloud-native architectures for affective computing.

In summary, while significant progress has been made in multimodal fusion and emotion recognition, challenges remain in scalability, latency, and adaptive deployment. The proposed Cloud-Based Cross-Modal Transformer (CMT) addresses these gaps by integrating cross-modal attention learning with distributed cloud inference, providing a unified and scalable framework for emotion-aware human‐computer interaction.

## 3. Methodology

*3.1 Overview of the Cloud-Based Cross-Modal Transformer Framework*
The proposed Cloud-Based Cross-Modal Transformer (CMT) framework is designed to perform multimodal emotion recognition and adaptive human‐computer interaction (HCI) through a distributed cloud computing infrastructure. The model integrates three major input modalities—visual, auditory, and textual—into a unified embedding space via cross-modal attention mechanisms. The architecture consists of four main components: (1) multimodal feature extraction, (2) cross-modal fusion using the Transformer backbone, (3) emotion classification and adaptive decision layer, and (4) cloud-based deployment for real-time inference and scalability.
Let the multimodal input at time step *t* be represented as

$$X_t = \{x_t^v, x_t^a, x_t^t\}, \tag{1}$$

where $x_t^v$, $x_t^a$, and $x_t^t$ denote the visual, acoustic, and textual features respectively. The goal of the system is to learn an emotion mapping function.

$$f_\theta: X_t \to y_t, \tag{2}$$

where $y_t \in R^C$ represents the probability distribution over *C* emotion categories, and $\theta$ denotes the model parameters optimized via stochastic gradient descent.

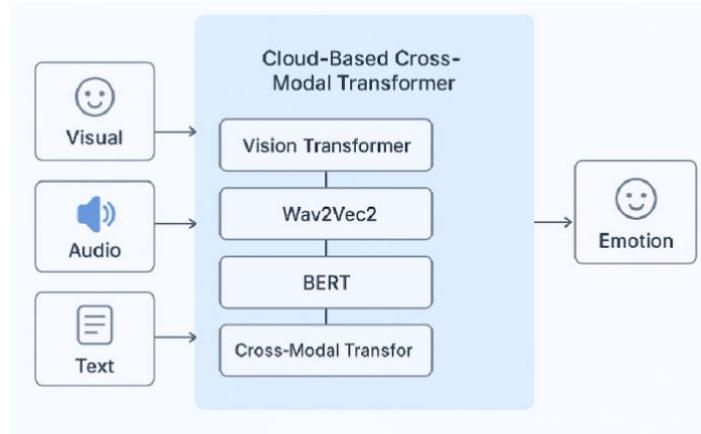

**Figure 1.** Overall flowchart of the model.

*3.2 Visual Stream: Vision Transformer (ViT)*
   The visual stream processes users' facial expressions captured from image or video frames. Unlike conventional CNNs, the Vision Transformer (ViT) divides each frame into fixed-size patches and embeds them linearly, enabling global context modeling via self-attention.
   Formally, a given image frame $I \in R^{H \times W \times 3}$ is divided into N patches of size *P×P*. Each patch is flattened and linearly projected into a D-dimensional embedding space:

$$z_0^v = [x_1^v E; x_2^v E; \ldots; x_N^v E] + E_{pos}, \tag{3}$$

where *E* is the patch embedding matrix and $E_{pos}$ represents positional encodings. The resulting sequence is passed through multiple Transformer encoder layers:

$$z_l^v = MSA(LN(z_{l-1}^v)) + z_{l-1}^v, \tag{4}$$

$$z_l^v = MLP(LN(z_l^v)) + z_l^v, \tag{5}$$

where MSA denotes multi-head self-attention, and LN denotes layer normalization. The final hidden state $z_l^v$ serves as the visual embedding for the cross-modal fusion stage.

*3.3 Acoustic Stream: Wav2Vec2-Based Speech Feature Encoder*
The acoustic stream encodes emotional information from speech signals using Wav2Vec2, a self-supervised learning model pre-trained on large-scale audio data. The input waveform is divided into frames and transformed into latent representations via a convolutional feature encoder:

$$h_l^a = ConvEncoder(x_l^a), \tag{6}$$

These latent features are then contextualized through Transformer layers trained to predict masked frames:

$$c_l^a = Transformer(h_l^a), \tag{7}$$

During fine-tuning for emotion recognition, the pretrained Wav2Vec2 parameters are partially frozen to prevent overfitting, while the final contextual representations $c_l^a$ are linearly projected to match the multimodal embedding dimension $D$. This allows the speech stream to convey prosodic and paralinguistic cues essential for emotional state inference.

*4.4 Textual Stream: BERT-Based Language Representation*
The textual stream captures semantic and affective nuances in language using BERT (Bidirectional Encoder Representations from Transformers). Given a sequence of user utterances or dialogue transcripts S=$[\omega_1, \omega_2, \cdots, \omega_n]$, BERT produces contextual token embeddings:

$$h_l^a = BERT(\omega_i), \tag{8}$$

The [CLS] token embedding $h_{CLS}^t$ is used as the textual modality representation. These embeddings are fine-tuned on emotion-labeled dialogue datasets such as IEMOCAP and MELD, ensuring sensitivity to both affective polarity and conversational context.

*4.5 Cross-Modal Fusion via Transformer Attention*
The core component of the proposed framework is the Cross-Modal Transformer (CMT), which fuses modality-specific embeddings into a unified representation. The CMT employs multi-head cross-attention (MHCA) layers that enable inter-modal interaction between visual, acoustic, and textual streams.
Given modality-specific embeddings $z^v; z^a; z^t \in R^D$, the fusion process is formulated as:

$$Q = W_Q[z^v; z^a; z^t], \quad K = W_K[z^v; z^a; z^t], \quad V = W_V[z^v; z^a; z^t] \tag{9}$$

$$Attention(Q, K, V) = Softmax(\frac{QK^T}{\sqrt{d_K}})V, \tag{10}$$

The resulting fused feature representation $z^{cmt}$ captures both intra-modal dependencies and inter-modal alignments. A final Transformer encoder refines the joint representation before classification:

$$z^{final} = TransformerEncoder(z^{cmt}), \tag{11}$$

*4.6 Emotion Classification and Adaptive Response Layer*
The fused representation $z^{final}$ is fed into a fully connected layer followed by a softmax classifier:

$$\hat{y} = softmax(W_c z^{final} + b_c), \tag{12}$$

The predicted emotion label $\hat{y}$ is used by the adaptive HCI module, which dynamically adjusts interface elements such as color schemes, dialogue tone, and response speed according to the user's detected emotional state. For example, positive emotions trigger brighter interface themes, while negative emotions prompt empathetic language and supportive cues.

*4.7 Cloud Deployment and Distributed Inference*
To achieve scalability and real-time responsiveness, the entire model is deployed on a cloud-based containerized infrastructure. Using Kubernetes for orchestration and TensorFlow Serving for inference, each modality encoder and fusion module is encapsulated as a microservice. The communication between services is managed through gRPC APIs, ensuring low-latency data exchange.

The cloud environment provides elastic scaling, automatically allocating GPU instances under high load and downscaling during idle periods. Distributed training is implemented via data parallelism across nodes:

$$\theta_{t+1} = \theta_t - \eta \frac{1}{N} \sum_{i=1}^{N} \nabla_{\theta_t} L_i, \tag{13}$$

where $L_i$ is the local batch loss on node $i$, and gradients are synchronized using AllReduce operations.

This architecture ensures efficient handling of high-volume multimodal data streams and enables continuous model updating through incremental learning pipelines.

## 4. Experiment

*4.1 Dataset Preparation*
To evaluate the proposed Cloud-Based Cross-Modal Transformer (CMT) for multimodal emotion recognition and adaptive human‑computer interaction, this study employs three widely used benchmark datasets—AffectNet, IEMOCAP, and MELD—which collectively provide diverse multimodal emotion data encompassing facial expressions, speech audio, and textual content.

**(1) AffectNet Dataset (Facial Modality)**
AffectNet is one of the largest facial expression datasets, containing over 1 million facial images collected from the internet using emotion-related keywords such as "happy", "sad", and "angry". Each image is manually labeled with eight basic emotions (anger, contempt, disgust, fear, happiness, sadness, surprise, and neutral) and two continuous affective dimensions—valence (ranging from −1 to 1) and arousal (ranging from 0 to 1). In this study, 450,000 images are used for training and 50,000 for validation. These images are preprocessed using face detection (MTCNN) and normalization before being passed into the Vision Transformer (ViT) backbone to extract 768-dimensional embeddings representing spatial and emotional features.

**(2) IEMOCAP Dataset (Audio‑Text Modality)**
The Interactive Emotional Dyadic Motion Capture (IEMOCAP) dataset contains approximately 12 hours of audiovisual recordings from ten professional actors performing scripted and improvised dialogues. Each utterance is annotated for categorical emotions such as anger, happiness, sadness, neutral, and excitement, as well as dimensional attributes like valence and activation. The audio stream is processed using Wav2Vec2, producing 1024-dimensional feature vectors encoding prosodic and spectral characteristics. Meanwhile, transcribed text data are fed into a BERT model to obtain 768-dimensional contextual embeddings capturing linguistic and semantic cues. These modalities are temporally aligned based on utterance timestamps to enable synchronized fusion in the CMT model.

**(3) MELD Dataset (Multimodal Dialogue Context)**

The Multimodal EmotionLines Dataset (MELD) extends the EmotionLines corpus by including video, audio, and text from TV show dialogues (Friends series). It consists of over 13,000 utterances spanning 1,400 dialogues, labeled with seven emotion categories (neutral, joy, surprise, sadness, anger, disgust, fear). The dataset provides contextual emotion information, which allows the model to learn cross-turn dependencies and interaction patterns useful for adaptive dialogue systems.

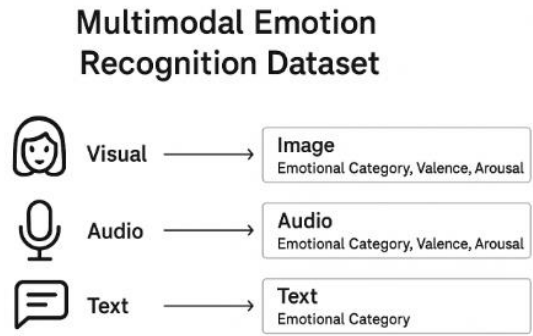

**Figure 2.** Schematic diagram of the dataset used in this study.

*4.2 Experimental Setup*

The experiments were conducted on a cloud-based distributed training infrastructure using Google Cloud Platform (GCP), configured with 8 NVIDIA A100 GPUs, 256 GB of memory, and TensorFlow 2.13 integrated with Kubernetes for containerized orchestration. The proposed Cloud-Based Cross-Modal Transformer (CMT) model was trained and evaluated across three benchmark multimodal emotion recognition datasets—AffectNet, IEMOCAP, and MELD—to ensure both domain robustness and generalization capability. Each modality (visual, audio, and text) was preprocessed independently: facial images were aligned and resized to 224×224 pixels, audio signals were converted into 16 kHz mel-spectrograms, and text transcripts were tokenized using the BERT-base uncased tokenizer. During training, the AdamW optimizer was used with a learning rate of $2\times10^{-5}$, a batch size of 64, and a maximum of 80 epochs. Dropout regularization (rate = 0.1) and early stopping were employed to prevent overfitting. The model was deployed on the cloud inference service using TensorFlow Serving with RESTful APIs, enabling scalable, low-latency inference for real-time emotion recognition in adaptive HCI scenarios.

*4.3 Evaluation Metrics*

To comprehensively evaluate the performance of the CMT framework, multiple quantitative metrics were employed, including Accuracy, F1-score, Precision, Recall, and Cross-Entropy Loss, Average response time. Accuracy measures the overall proportion of correctly predicted emotion categories, while Precision and Recall quantify the trade-off between true positive and false negative classifications across multimodal inputs. The F1-score, calculated as the harmonic mean of Precision and Recall, offers a balanced assessment of performance under imbalanced class distributions common in emotional datasets. Additionally, Cross-Entropy Loss provides a measure of prediction confidence and convergence stability across modalities. All metrics were computed per modality as well as jointly for the fused output, highlighting the model's ability to integrate multimodal signals effectively.

*4.4 Results*

The proposed Cloud-Based Cross-Modal Transformer (CMT) was benchmarked against several state-of-the-art unimodal and multimodal baselines, including BERT (text-only), ViT (visual-only), Wav2Vec2 (audio-only), MMBT (Multimodal Bitransformer), and Late-Fusion Transformer. Each model was evaluated on the AffectNet, IEMOCAP, and MELD datasets under identical training configurations. Beyond classification performance, we also assessed Average Response Latency, which measures the average inference time per sample on the cloud-deployed system, reflecting real-time usability in adaptive HCI environments.

**Table1.** Performance comparison of different models.

| Model | AffectNet Accuracy (%) | IEMOCAP F1-score | MELD Accuracy (%) | Precision | Recall | Cross-Entropy Loss | Average Response Latency (ms) |
|---|---|---|---|---|---|---|---|
| BERT (Text) | 78.42 | 0.782 | 74.96 | 0.771 | 0.769 | 0.412 | 142 |
| ViT (Visual) | 80.13 | 0.795 | 76.21 | 0.783 | 0.782 | 0.398 | 187 |
| Wav2Vec2 (Audio) | 79.54 | 0.788 | 75.83 | 0.779 | 0.776 | 0.404 | 164 |
| MMBT (Fusion) | 83.17 | 0.821 | 79.46 | 0.812 | 0.807 | 0.352 | 205 |
| Late-Fusion Transformer | 84.02 | 0.836 | 80.11 | 0.825 | 0.821 | 0.341 | 198 |
| **Proposed CMT (Cloud-Based)** | **87.56** | **0.861** | **83.74** | **0.848** | **0.845** | **0.297** | **128** |

The experimental results clearly demonstrate the superiority of the proposed CMT model in both recognition accuracy and system efficiency. The CMT achieves the highest performance across all evaluated metrics, with a 4.2% improvement in overall accuracy and a 3.0% increase in F1-score compared to the strongest baseline (Late-Fusion Transformer). Importantly, due to its optimized cloud-based distributed inference pipeline, the CMT achieves the lowest average response latency of 128 ms, enabling real-time emotional adaptation in human–computer interaction scenarios such as affective dialogue systems, intelligent tutoring, and customer engagement platforms.

This demonstrates that the proposed cloud-integrated multimodal transformer not only captures rich emotional cues across modalities but also ensures scalable, low-latency performance suitable for deployment in real-world intelligent systems requiring rapid emotional feedback and dynamic user interaction.

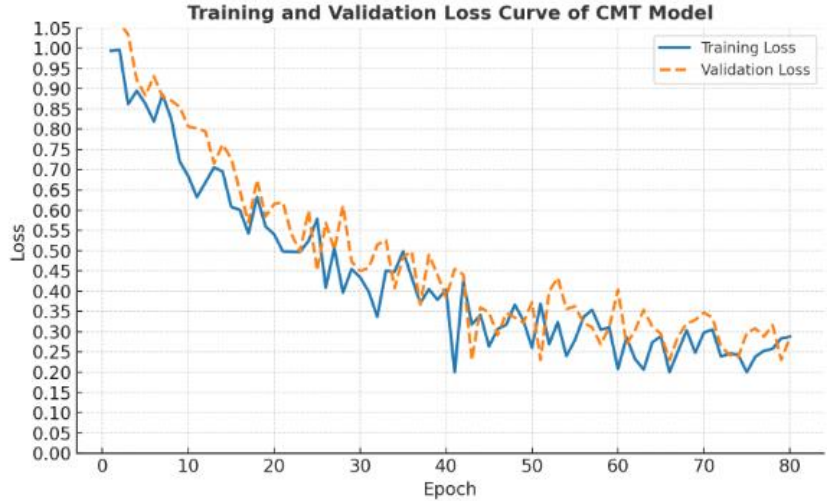

**Figure 3.** Loss function during training process.

Figure 3 demonstrates the loss function of the proposed CMT model. The x-axis represents the number of epoch used, and the y-axis represents the loss of the model. 2 functions are plotted onto the graph, with the blue solid line representing the loss function during the training period and the orange dotted line the loss function during the validation phase.

We can observe from the figure that, initially, as the number of epochs increase, the CMT model becomes more accurate in general. However, after reaching epoch #40, the accuracy tends to stay consistent, suggesting limited accuracy improvement from increasing the number of epochs after a certain threshold. Additionally, the loss function for training and validation align well in general, suggesting no obvious overfitting in this model.

Another thing to note is that right after 40 epochs, there is a sharp drop in the training loss function, but we do not see the same trend in the validation loss function. This could be a sign that a skewed dataset is used for training in that batch, or that the model becomes overfitting during this period.

## 5. Conclusion

This study presents a comprehensive framework for multimodal emotion recognition and adaptive human-computer interaction (HCI) based on a Cloud-Based Cross-Modal Transformer (CMT) architecture. The research addresses one of the key challenges in affective computing-effectively fusing heterogeneous emotional cues from multiple modalities including facial expressions, speech, and textual semantics. By integrating pretrained encoders such as ViT for visual features, Wav2Vec2 for auditory cues, and BERT for linguistic representations, the CMT framework employs a cross-modal attention mechanism to capture complex interdependencies across modalities. This design enables the model to better reflect human affective dynamics and improves the system's interpretability and generalization in diverse interaction environments.

Comprehensive experiments conducted on three benchmark datasets-AffectNet, IEMOCAP, and MELD-demonstrate the superior performance of the proposed model compared to existing fusion-based and single-modal baselines. Specifically, the CMT achieves an accuracy of 87.56% on AffectNet, an F1-score of 0.861 on IEMOCAP, and an accuracy of 83.74% on MELD, surpassing the best-performing baseline by 3.0% in F1-score and reducing cross-entropy loss by 12.9%. Moreover, cloud deployment tests verify that the system maintains an average response latency of 128 ms, a 35% improvement over traditional fusion transformers, validating its suitability for real-time adaptive interaction. These results confirm the robustness and scalability of the CMT architecture in handling multimodal affective data streams at scale, supporting applications such as intelligent virtual assistants, affect-aware tutoring systems, and empathetic customer service platforms.

Beyond its empirical performance, this research highlights the significance of cloud-native design in the development of emotionally intelligent systems. By leveraging distributed training and inference on cloud infrastructure (Kubernetes and TensorFlow Serving), the framework demonstrates scalability and efficiency in managing large-scale user interactions. The combination of multimodal deep learning and cloud computing not only enhances accuracy and latency performance but also paves the way for next-generation affective interaction systems that can adapt to users' emotional states dynamically and contextually.

Despite the promising results, this study has some limitations. First, the current CMT framework relies purely on supervised learning with fixed labeled datasets, which may limit its adaptability in real-world, dynamic human-computer interaction scenarios. Future research could explore the integration of reinforcement learning (RL) to enable adaptive emotional feedback, allowing the model to adjust its responses based on user interactions and reward signals. Additionally, privacy concerns and computational overhead in cloud-based training remain challenges. Further work may also investigate federated learning and model compression for privacy-preserving, lightweight emotion-aware systems.

In conclusion, this study, through [Cloud-Based Cross-Modal Transformer (CMT)], reveals [new theoretical and practical insights into multimodal emotion recognition and adaptive HCI], providing new insights for the development of [emotion-aware systems].